\documentclass[runningheads]{llncs}

\usepackage{amsmath}
\usepackage{amssymb}
\usepackage{amsfonts}

\usepackage[table,xcdraw]{xcolor}
\usepackage{multirow}
\usepackage{booktabs}
\usepackage{graphicx}
\usepackage{wrapfig}
\usepackage{subcaption}
\usepackage{booktabs}
\usepackage{bigstrut}
\usepackage{tikz}
\usetikzlibrary{automata,arrows.meta,positioning,shapes,fit,backgrounds,calc}

\usepackage{algorithm}

\usepackage{amsthm}
\usepackage[noend]{algpseudocode}
\usepackage[subtle]{savetrees}

\usepackage{hyperref}
\usepackage{cleveref}

\usepackage{xspace}

\newcommand{\differ}{\texttt{DIFF-ERO}\xspace}

\usepackage{fontspec}  

\begin{document}

\title{\differ: A Conformance-Aware Loss for \\Deep Learning in Process Mining}

\titlerunning{\differ: Differentiable Entropic Process Loss}

\author{}
\institute{}
\author{Johannes~De~Smedt\inst{1}\orcidID{0000-0003-0389-0275} 
\institute{Information Systems Engineering Research Group (LIRIS), KU Leuven\\
Naamsestraat 69, 3000 Leuven, Belgium\\
 \email{johannes.desmedt@kuleuven.be}}
 }
\author{Johannes~De~Smedt\inst{1} \and Jari~Peeperkorn\inst{1} \and Artem~Polyvyanyy\inst{2} \and Jochen~De~Weerdt\inst{1}}

\institute{
KU Leuven, Leuven, Belgium\\
\email{\{johannes.desmedt;jari.peeperkorn;jochen.deweerdt\}@kuleuven.be}\\
\and
The University of Melbourne, Parkville, VIC 3010, Australia\\
\email{artem.polyvyanyy@unimelb.edu.au}
}
\authorrunning{De Smedt et al.}

\maketitle

\begin{abstract}
Deep learning has driven many recent advances in process analytics, especially for predictive and prescriptive monitoring. However, standard objectives such as cross-entropy optimize local next-step likelihoods and only implicitly capture control-flow structure. As a result, models can achieve high token-level accuracy while permitting imprecise global behaviour.
We introduce \differ, a conformance-aware loss function for deep learning models on process data. \differ is a differentiable formulation of entropy-based stochastic conformance that incorporates control-flow information during training. Our approach constructs batch-level stochastic transition matrices with soft edge memberships, allowing structural precision and recall signals to directly inform backpropagation.
The loss is model-agnostic and can be applied whenever the final representation parametrizes stochastic transitions.
We instantiate \differ in transformer encoder–decoder pipelines for next-activity prediction and use it jointly with cross-entropy to analyse its theoretical components with respect to convergence. 
Across benchmarks comparing other loss functions and targets, \differ shows improved predictive performance where structure matters most while maintaining parity elsewhere. At the same time, the learned stochastic automaton converges towards the structural ground truth, indicating that the network internalizes process model structure. 



\end{abstract}

\section{Introduction}\label{sec:intro}
Deep neural models, especially recurrent and transformer-based architectures, have advanced predictive and prescriptive process monitoring.  Since the early use of LSTMs~\cite{tax2017predictive}, researchers have applied deep learning to outcome prediction, next-activity and suffix prediction, and remaining-time estimation. Graph neural networks and transformers have further broadened the toolbox~\cite{kappel2024attention,wuyts2024sutran}. In prescriptive settings, LSTMs enable deep reinforcement learning to optimize process performance~\cite{de2025procause}, and even process discovery can be cast as a deep learning task~\cite{sommers2021process}.

Despite this progress, most approaches solely optimize standard objectives such as cross-entropy or squared error. Cross-entropy is effective for local prediction targets, yet it is precision-focused, susceptible to class imbalance~\cite{terven2025comprehensive}, and unable to capture structural information such as correlations in the output space. This has led researchers in image and text classification to formulate alternative losses~\cite{graber2018deep,mnih2012conditional}.
In event logs, cross-entropy loss does not explicitly represent process structures such as loops, long-distance dependencies, concurrency, and choice. 
As a result, models can achieve high next-step accuracy while permitting imprecise global behaviour. Deep learning models may then fail to generalize to unseen, but correct, control-flow behaviour~\cite{peeperkorn2023rnn,peeperkorn2024validation}. 

Recent efforts encode control flow into features or tensors~\cite{donadello2024conformance}, or impose control-flow constraints on generated outputs via offline conformance calculations during inference~\cite{buliga2025guiding,stevens2025generating}. However, these strategies do not inform the learning signal during gradient calculations and can yield high next-event accuracy without guarantees of process model conformance. We therefore seek a training objective that makes control-flow conformance part of the signal received during learning, encouraging structural fidelity in addition to local accuracy.

We address this gap with a loss function specifically designed for deep learning in process analytics. Building on entropic relevance, a stochastic conformance metric, we develop a DIFFerentiable EntRopic Process LOss \differ that brings control-flow conformance information directly into training. \differ complements the recall-oriented behaviour encouraged by cross-entropy with a precision-oriented signal rooted in process conformance. We demonstrate the objective in a multi-modal transformer encoder–decoder that decodes an automaton representation of traces and compares it against a ground-truth control-flow model.

Our contributions are as follows:
\begin{itemize}
  \item \textbf{Differentiable entropic process loss.} A neural-compatible formulation of entropic conformance with soft edge memberships in a stochastic automaton and a batch-level fallback term, yielding a fully differentiable conformance-aware loss.
  \item \textbf{Batchable conformance supervision.} A scheme to construct ground-truth stochastic transition information from (mini-)batches or the full training log to supervise structural fit during backpropagation.
  \item \textbf{Theoretical analysis.} The proposed loss functions are shown to converge to desirable properties. 
  \item \textbf{Integration with predictive tasks.} A multi-task setup that combines \differ with cross-entropy for transformer-based next-activity 
  prediction, enabling both local accuracy and global structural fidelity.
  \item \textbf{Empirical study.} An evaluation on 
  benchmark transformer architectures for next-activity 
  prediction that shows convergence in terms of learned process model structure and identifies where \differ improves predictive accuracy. 
\end{itemize}

The paper is structured as follows. Section~\ref{sec:2background} introduces background and motivation. Section~\ref{sec:preliminaries} presents the preliminaries, and Section~\ref{sec:method} introduces the \differ loss and its differentiable formulation. Section~\ref{sec5:empirical} describes the experimental setup and results, Section~\ref{sec:discussion} discusses limitations, and Section~\ref{sec:conclusion} concludes.

\section{Background and Motivation}\label{sec:2background}

Conformance checking quantifies how well a process model explains an event log \cite{carmona2018conformance}. Early works, such as token-based replay, evaluate fitness and precision by replaying traces on a Petri net and counting produced and remaining tokens. Alignment-based conformance finds the lowest-cost sequences of synchronous and model/log moves so that each trace fits the model, typically solved with search or ILP. Behavioural precision and recall measure, respectively, how much model behaviour appears in the log and how much observed behaviour is allowed by the model.
Recent conformance checking approaches shifted towards comparing \emph{distributions} or the stochasticity of behaviour. Earth Mover's Distance (EMD) measures the cost of transforming one probability distribution over traces into another, but becomes expensive over large trace spaces and long prefixes \cite{leemans2019earth}. The entropic relevance (ER) family models behaviour with a stochastic automaton and uses entropy to quantify bits needed to replay a trace, with a regularization that discourages overly permissive models \cite{alkhammash2022entropic}. Standard formulations of these metrics, however, are not directly differentiable as they rely on discrete calculations.

Deep learning approaches have surfaced for various predictive and prescriptive process analytics tasks. 
Neural sequence models such as Long Short-term Memory networks (LSTMs) and Transformers are widely used for next-activity prediction, remaining time, and suffix prediction.
These black box models offer no guarantees for capturing control flow information directly \cite{peeperkorn2023rnn}.
Hence, recent approaches have shifted towards injecting conformance information into these models as additional input or auxiliary signals, 
for example, by providing a process model alongside the prefix \cite{chiorrini2021exploiting,lischka2025directly}, or via fuzzy/tensor encodings of control-flow constraints \cite{donadello2024conformance,umili2023grounding}. These strategies enrich representations at the input or provide an offline-calculated conformance metric \cite{buliga2025guiding}, but do not by themselves define a training objective based on conformance.

Beyond process mining, losses tailored to evaluation criteria for sequential data include minimum-risk and calibration objectives in NLP \cite{edunov2017classical,zhao2022calibrating}, and Soft-DTW for time series \cite{cuturi2017soft}. These are differentiable and align gradients with task semantics, but they do not account for process-model conformance. Additionally, contrastive learning and learning-to-rank losses have emerged in the context of models working over sequences such as text \cite{chen2009ranking,xie2022contrastive} to improve robustness and generalization capabilities of predictive models. E.g., \cite{xie2022contrastive} breaks up next token tasks in transformer networks into a double loss component to provide contrastive/negative samples to the loss function. 
In the further neuro-symbolic learning domain, various recent works have proposed differentiable loss functions to inform the training about, e.g., temporal relations \cite{mezini5420992neuro,xu2022don}. 

Given that conformance checking techniques such as token replay, alignments, or sequence constraints are discrete in nature, they are not compatible with the continuous, differentiable backpropagation-based learning of neural networks.
The neuro-symbolic approaches which offer the potential to input conformance-based or constraint information into a deep learning model either use offline pre-calculated conformance results \cite{buliga2025guiding,stevens2025generating}, or use discrete approximations using Gumbel-softmax \cite{mezini5420992neuro,umili2023grounding} which typically rely on expensive sampling procedures to achieve these approximations \cite{paulus2020gradient}.
Furthermore, these approaches typically require either prior domain knowledge \cite{donadello2023knowledge,umili2023grounding} or a dedicated discovery step except for \cite{mezini5420992neuro,stevens2025generating}.

In the next sections, we introduce a natively differentiable objective derived from the natively stochastic, continuous domain of entropic relevance that connects these lines of work by bringing process-model conformance directly into neural training.

\section{Preliminaries}\label{sec:preliminaries}
We recall basic concepts on event logs, entropic relevance, and predictive model training.

\subsection{Event Logs}
\label{sec:event:logs}
Let $\mathcal{U}_A$ and $\mathcal{U}_C$ be the universes of activities and case identifiers, respectively.
An \emph{event log} $E$ is a finite set of events. 
An \emph{event} $e \in E$ is a triple $(a,t,c)$, where $a \in A\subseteq \mathcal{U}_A$ is the activity that triggered the event, $t \in \mathbb{R}$ is the timestamp at which the event was triggered or recorded, and $c \in C\subseteq \mathcal{U}_C$ is the identifier of the case in which activity $a$ was executed (e.g., the case of handling a specific loan application). 
We assume that no two events share the same timestamp.
By $e^a$, $e^t$, and $e^c$, we denote the activity, timestamp, and the case identifier of event $e$, respectively.
A \emph{trace} of case $c$ is denoted by $\sigma_c$, and is the sequence $\langle e_1, \ldots, e_n \rangle$, $n \in \mathbb{N}$, of all events in $E$ with case identifier $c$ ordered by their timestamps, i.e., 
$\forall\, i \in [1 \, .. \, n] : e_i^c=c$, 
$\forall\, i,j \in [1 \, .. \, n], i < j : e^t_i < e^t_j$, and 
$\forall e \in E : (e^c=c) \implies (\exists\, i \in [1 \,..\, n] : e_i = e)$.
For ease of notation, we also use $\sigma \in E$ to denote a trace whose events are contained in $E$.

For instance, consider event log \(E=\{e_1,e_2,e_3,e_4,e_5,e_6\}\) over activities \(\mathcal{A}=\{\texttt{start},\texttt{review},\texttt{approve},\texttt{complete}\}\) with events as follows: \[
  \begin{aligned}
    e_1 &= (\texttt{start},\,1.00,\,c_1),&
    e_2 &= (\texttt{review},\,1.05,\,c_1),\\
    e_3 &= (\texttt{approve},\,1.10,\,c_1),&
    e_4 &= (\texttt{complete},\,1.15,\,c_1),\\
    e_5 &= (\texttt{start},\,2.00,\,c_2),&
    e_6 &= (\texttt{complete},\,2.05,\,c_2).
  \end{aligned}
\]
This event log describes two traces:
\[
  \sigma_{c_1}=\langle e_1,e_2,e_3,e_4\rangle,
  \qquad
  \sigma_{c_2}=\langle e_5,e_6\rangle.
\]
Throughout this paper, we assume that the first and last events of each trace are triggered by special \texttt{start} and \texttt{complete} activities, respectively.

\subsection{Entropic Relevance}
Entropic relevance is a stochastic conformance checking metric that quantifies how well a stochastic process model, that is, a process model that describes the probabilities of the decisions in the process, discovered from a given event log, explains the distribution of traces the system that generated the event log executes~\cite{polyvyanyy2020entropic,alkhammash2022entropic}.
The entropic relevance of a stochastic process model is the average number of bits used to compress a trace from the event log using the probabilities of traces described by the model.
Fewer bits indicate that the model better captures the trace probabilities of the system.

A Stochastic Deterministic Finite Automaton (SDFA) $\mathcal{A}$ is a tuple $(S,A,\delta,\delta_p,s_0)$, where 
$S$ is a finite set of states, 
$A$ is a set of activities, 
$\delta: S \times A\to S$ is the transition function 
annotated with probabilities by 
function $\delta_p:S \times A\to [0,1]$, such that for each state $s \in S$ it holds that $\sum_{a \in A} \delta_p(s,a) \leq 1$, and 
$s_0$ is the initial state.

An SDFA $\mathcal{A}$ induces a stochastic language $L_\mathcal{A}$.
In this language, the probability of a trace is defined as the product of the probabilities of the transitions 
along the unique execution in the automaton induced by the trace, multiplied by the termination probability of the state reached after replaying the trace. 
The termination probability of a state $s \in S$ is given by 
$1 - \sum_{a \in A} \delta_p(s,a)$.

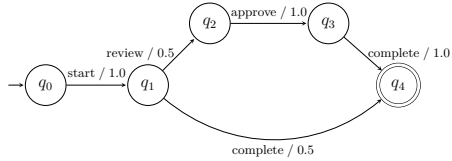
\begin{figure}
\vspace{-1mm}
\begin{center}
\resizebox{.5\textwidth}{!}{
\begin{tikzpicture}[
    shorten >=1pt,
    node distance=1.5cm, 
    initial text=,
    state/.style={circle, draw, minimum size=1cm, font=\large},
    accepting/.style={double, double distance=2pt},
    >=stealth 
]
    \node[state, initial] (q0) {$q_0$};
    \node[state] (q1) [right=of q0] {$q_1$};
    \node[state] (q2) [above right=0.8cm and 0.8cm of q1] {$q_2$}; 
    \node[state] (q3) [right=1.9cm of q2] {$q_3$};
    \node[state, accepting] (q4) [below right=0.8cm and 1cm of q3] {$q_4$};
    \draw[->] (q0) edge node[above] {start / 1.0} (q1);
    \draw[->] (q1) edge node[left] {review / 0.5} (q2);
    \draw[->] (q1) edge[bend right=40] node[below] {complete / 0.5} (q4);
    \draw[->] (q2) edge node[above] {approve / 1.0} (q3);
    \draw[->] (q3) edge node[right] {complete / 1.0} (q4);
\end{tikzpicture}
}
\end{center}
\vspace{-3mm}
\caption{An example SDFA.}
\label{fig:sdfa_example}
\vspace{-2mm}
\end{figure}

Consider SDFA $\mathcal{A}$ in \Cref{fig:sdfa_example}.
It holds that 
$L_\mathcal{A}(\sigma_{c_1}) = \Pr(\sigma_{c_1} \mid \mathcal{A}) = 1.0 \times 0.5 \times 1.0 \times 1.0  \times 1.0 = 0.5$ and 
$L_\mathcal{A}(\sigma_{c_2}) = \Pr(\sigma_{c_2}\mid \mathcal{A}) = 1.0 \times 0.5  \times 1.0 = 0.5$.

By $\Pr(\sigma \mid E)$, we denote the relative likelihood of trace $\sigma$ in event log $E$
, that is, the empirical probability of observing trace $\sigma$ among all traces in $E$.
By $E_\mathcal{A}$, we denote the sublog of $E$ obtained by retaining all traces $\sigma$ such that $L_\mathcal{A}(\sigma) > 0$, that is, all the traces in $E$ that are described in $\mathcal{A}$.

Then, by $\rho(E,\mathcal{A})$, we denote the probability that a trace in $E$ is described in $\mathcal{A}$:
\begin{equation}\label{eq:rho}
\rho(E,\mathcal{A})=\sum_{\sigma \in E_\mathcal{A}} \Pr(\sigma \mid E) \, .
\end{equation}
\noindent
For example,
$\rho(E,\mathcal{A})
    = \Pr(\sigma_{c_1} \mid E) + \Pr(\sigma_{c_2} \mid E)
    = \frac12 + \frac12
    = 1$,
where $E$ and $\mathcal{A}$ are the example event log and automaton in \Cref{sec:event:logs} and \Cref{fig:sdfa_example}, respectively.

By $\left\|E\right\|$, we denote the total number of traces recorded in $E$.
Then, \emph{entropic relevance} of SDFA $\mathcal{A}$ to $E$ is denoted by $\mathit{ER}(E,\mathcal{A})$ and defined below:
\begin{equation}\label{eq:er}
\begin{split}
\mathit{ER}(E,\mathcal{A})=H_0(\rho(E,\mathcal{A}))+ \frac{1}{\left\|E\right\|} \sum_{\sigma \in E}\mathit{cost}_{\mathit{bits}}(\sigma,E,\mathcal{A}) \, .
\end{split}
\end{equation}
\begin{equation}\label{eq:cost}
\mathit{cost}_{\mathit{bits}}(\sigma,E,\mathcal{A}) = \begin{cases}
-\log_2 L_\mathcal{A}(\sigma) & \text{if } L_\mathcal{A}(\sigma) > 0 \, , \\
\text{bits}(\sigma,E,\mathcal{A}) & \text{otherwise} \, .
\end{cases}
\end{equation}
\begin{equation}\label{eq:h0}
\begin{split}
H_0(p)=-p\;\;\log_2(p) - (1-p)\;\;\log_2 (1-p) \; , \;\;\;\; p \in [0,1]\; .
\end{split}
\end{equation}

Function $\text{bits}(\sigma,E,\mathcal{A})$ implements a background procedure that penalizes traces not compatible with the model.
For example, the uniform background coding model is calculated as $(1+|\sigma|)\log_2(1+|A|)$, see~\cite{polyvyanyy2020entropic} for details. 

For the running example, it holds that $\rho(E,\mathcal A) = 1$, resulting in $H_0\!\bigl(\rho(E,\mathcal A)\bigr)
  = 0$; taking $H_0(0) = H_0(1) = 0$ by definition.
In addition, it holds that:
\[
cost_{\mathit{bits}}\bigl(\sigma_{c_1},E,\mathcal A\bigr)
= cost_{\mathit{bits}}\bigl(\sigma_{c_2},E,\mathcal A\bigr)
   = -\log_2 0.5 = 1\;\text{bit}.\]
Hence, it holds that $\mathit{ER}(E,\mathcal A)
  = 0
  + \frac{1}{2}\bigl(1+1\bigr)
  = 1\;\text{bit}$.
For an example non-fitting trace $\sigma_{\mathit{nf}} = \langle e_1,e_2,e_6\rangle$, with $|A|=4$, the penalty-term is engaged as below:
\[
\text{bits}(\sigma_{\mathit{nf}},E,\mathcal A)
  =(1+|\sigma_{\text{nf}}|)\log_2(1+|A|)
  = 4\,\log_2 5
  \approx 9.29\;\text{bits}.
\]

\noindent
If this trace is included in $E$ to obtain event log $E'$, 
then it holds that
$\mathit{ER}(E',\mathcal A) \approx 0.92 + \frac{1}{3} \bigl(1+1+9.29\bigr) \approx 4.68$ bits.

\subsection{Predictive Model Training}
Predictive models, like next-activity or remaining-time prediction models, typically rely on prefixes of events~\cite{kappel2024attention}.
The prefix of event $e$ of length $l$ is the prefix of length $l \in \mathbb{N}$ of trace $\sigma_{e^c}$. 
We define the suffix of event $e$ analogously as the suffix of length $l$ of the trace defined by the case of event $e$.


Event prefixes and suffixes are often truncated and/or padded to ensure a consistent length for neural network models that use fixed input and output dimensions.
Therefore, by $\operatorname{pref}(e,l)$ and $\operatorname{suff}(e,l)$ we denote functions that return padded prefixes and suffixes, respectively, of event $e$ of length $l$.
\Cref{tab:prefixes:and:suffixes} shows padded prefixes and suffixes of length three for events from the example event log in \Cref{sec:event:logs}; the padding symbol is `$\bot$'.

\begin{table}[h!]
\vspace{-7mm}
\caption{Prefixes and suffixes (\(l=3\)) for the example event log from \Cref{sec:event:logs}.}
\smallskip
\centering
\resizebox{.7\textwidth}{!}{
\begin{tabular}{
  >{\centering\arraybackslash}p{1.8cm}  
  >{\centering\arraybackslash}p{2cm}    
  >{\centering\arraybackslash}p{2cm}    
  >{\centering\arraybackslash}p{2cm}    
  >{\centering\arraybackslash}p{2cm}    
}
\toprule
Event & Prefix & \(\operatorname{pref}(e_i,3)\) & Suffix & \(\operatorname{suff}(e_i,3)\) \\
\midrule
\(e_1\) & \(\langle\rangle\) & \(\langle \bot,\bot,\bot\rangle\) & \(\langle e_2,e_3,e_4\rangle\) & \(\langle e_2,e_3,e_4\rangle\) \\
\(e_2\) & \(\langle e_1\rangle\) & \(\langle \bot,\bot,e_1\rangle\) & \(\langle e_3,e_4\rangle\) & \(\langle e_3,e_4,\bot\rangle\) \\
\(e_3\) & \(\langle e_1,e_2\rangle\) & \(\langle \bot,e_1,e_2\rangle\) & \(\langle e_4\rangle\) & \(\langle e_4,\bot,\bot\rangle\) \\
\(e_4\) & \(\langle e_1,e_2,e_3\rangle\) & \(\langle e_1,e_2,e_3\rangle\) & \(\langle\rangle\) & \(\langle \bot,\bot,\bot\rangle\) \\
\(e_5\) & \(\langle\rangle\) & \(\langle \bot,\bot,\bot\rangle\) & \(\langle e_6\rangle\) & \(\langle e_6,\bot,\bot\rangle\) \\
\(e_6\) & \(\langle e_5\rangle\) & \(\langle \bot,\bot,e_5\rangle\) & \(\langle\rangle\) & \(\langle \bot,\bot,\bot\rangle\) \\
\bottomrule
\label{tab:prefixes:and:suffixes}
\end{tabular}
}
\end{table}

\vspace{-1.0cm}
\section{\differ: Differentiable Entropic Process Loss}\label{sec:method}
Section~\ref{sec:2background} outlined various conformance checking and process model featurisation approaches used in deep learning.
On the one hand, the use of discrete operations in conformance checking, such as aligning traces and models through ILP or $\in$-operators, impedes differentiable optimisation.
On the other hand, the use of Declare constraints~\cite{mezini5420992neuro,buliga2025guiding,stevens2025generating} or other behavioural profiles would again require a discrete one-hot encoding inside the loss function. 
\cite{mezini5420992neuro} provide a differentiable DeepDFA with approximations using Gumbel-softmax.
There is a trade-off between one the one hand having very expressive temporal imputation and tractability on the other hand as differentiable approximations relying on expensive sampling for discrete information which also do not estimate the joint distribution of several one-hot encoded variables as they do not natively account for correlations~\cite{paulus2020gradient}
, and multivariate solutions such as Gumbel-Sinkhorn can become expensive for larger feature sets~\cite{mena2018learning}.

We aim to develop a natively differentiable, neural network-compatible loss function that approximates the continuous distributions of stochastic process models, in analogy to how cross-entropy minimizes the Kullback-Leibler divergence in classification. 
We achieve this by first adapting the ideas of ER to the batch structure required for neural training and then making all terms differentiable so that the loss is compatible with backpropagation.

\subsection{Batch Calculations}\label{sec:batches}
Entropic relevance is calculated from the whole event log $E$.
However, in a neural network, the loss is typically calculated over batches (e.g., prefixes of events in $E$ with their corresponding targets), and gradients are backpropagated to inform tasks such as next-activity or outcome prediction.
We therefore adapt the ER calculations to obtain, for each batch $b$, a probability matrix $\mathit{PM}_b$ that plays the role of a ground-truth SDFA for that batch and share it across all samples or prefixes in $b$ (when calculating the loss for backpropagation). 
This ground-truth SDFA can then be compared with an estimated SDFA produced by the model.

To construct $\mathit{PM}_b$ we use directly-follows relations $DF_{\sigma}:A \times A\to \mathbb{N}$. 
For a sequence $\sigma$, i.e., 
$DF_{\sigma}(a_1,a_2)=|\{(e_1,e_2) \mid e^t_1 <e^t_2, \nexists e^t_1 < e^t_3 < e^t_2 \wedge e^a_1=a_1\wedge e^a_2=a_2, \forall e_1,e_2,e_3 \in \sigma\}|$.\footnote{Note that there are other representations from which an SDFA could be derived, e.g., stochastic Petri nets~\cite{leemans2024stochastic}.}
We can obtain $DF_{\sigma}$ for a set of sequences, e.g., all suffixes from all events in a batch, or even all suffixes from all events in a training log, by considering a subset of events $b \subseteq E_{train}\subset E$.
E.g., for a batch $b$ of events we can obtain the prefixes $\sigma_{pre}=\{\operatorname{pref}(e_i,l)\mid e_i \in b\}$ and their corresponding suffixes $\sigma_{suf}=\{\operatorname{suff}(e_i,l)\mid e_i \in b\}$ for a chosen $l$ so we can calculate 
$DF_{\sigma_{suf}}(a_1,a_2)=\Sigma_{\sigma \in \sigma_{suf}}DF_\sigma(a_1,a_2)$.
The use of directly-follows relations allows us to obtain a ground-truth probability matrix $PM_b \in [0,1]^{|A|\times |A|}$ with its elements $p_{a_1,a_2}=\frac{DF_b(a_1,a_2)}{\Sigma_{a_3} DF_b(a_1,a_3) + \epsilon}$ for batch $b$. Note that we include $\epsilon$ to avoid numerical instability in tensor calculations.

\subsection{Differentiable Calculations}

As explained in Section \ref{sec:2background}, conformance-based metrics such as ER rely on some form of replay. In the case of ER this is expressed in the form of $\sigma \in L_\mathcal{A}$ per equation \ref{eq:cost}. This discrete check impedes the formula from being used in a differentiable setup such as the backpropagation in neural networks. 

The first change towards a differentiable ER is to introduce a soft membership function over the edges of the original SDFA's transition probabilities $\delta_p$.
In a neural network, we obtain such a tensor from the output of the final hidden layer $h_O\in \mathbb{R}^{|B|\times |A| \times |A|}$ produced by, e.g., an LSTM or Transformer architecture ingesting prefix sequences, to derive $O$ from a softmax layer returning probabilities for the full transition matrix $O=softmax(h_O)$ with $softmax(x_i)=\frac{e^{x_i}}{\Sigma_{j} e^{x_j}}$. 
We refer to $O_b\in \mathbb{R}^{|A|\times|A|}$ as the batch-level representation of $\delta_p$ with a (near) 0 probability indicating the arc not being present in $\delta$.
$O_b$ can be used to calculate a batch-specific tensor $L_{PM_b}$ representing an output SDFA of the model to calculate the final difference with the batch-level ground-truth $PM_b$.
$O_b$ is globally normalized due to the softmax operator. An SDFA requires row-level normalization to ensure that for each state the outgoing probabilities sum to one.
Therefore, we apply an additional row-wise normalization $L_{PM_b}$ to obtain a valid conditional distribution.
$L_{PM_b}$ is calculated by 
\begin{equation}\label{eq:lpm}
L_{PM_{b}}(a_1,a_2)=\frac{O_b[a_1,a_2]}{\Sigma_{a_3\in A}O_b[a_1,a_3] + \epsilon} \in [0,1]^{|A|\times |A|}
\end{equation}
to transform the edge memberships into probabilities which sum to 1.

A second challenge is the $bits$-term from Equation \ref{eq:cost} being calculated based on the size of $t$ and $A$.
We propose a smoothed negative log-likelihood derived from $PM_b$, which signifies a form of fallback regularisation based on the ground truth.
If an edge is incorrectly predicted to (not) be part of the model, this term penalizes the neural network for such errors.
Other possible alternatives for this term include using the Gumbel-softmax trick to approximate the actual discrete values of $|t|$ and $|A|$, which might prove computationally expensive \cite{paulus2020gradient}, or differentiable placeholder terms learned with the model, which will mostly facilitate gradient calculations and have a similar purpose to our solution.
We obtain the fallback using standard cross-entropy to approximate $\rho(E,\mathcal{A})$ on batch-level as $\rho(E_b,PM_b)=\Sigma_{t\in E_b} Pr(t|E_b)$ by calculating
$-\log_2(PM_b)$.

Finally, we can rewrite Equation \ref{eq:cost} as the \differ~equation to measure the loss between ground-truth $PM_b$ and trained representation $O_b$ as
\begin{equation}\label{eq:differo}
\begin{split}
  \mathcal{L}_{DIFF-ERO}(E,PM_b,O_b)=  O_b\cdot (-\log_2 L_{PM_b}) + (1- O_b) \cdot  (-\log_2 PM_b) 
\end{split}
\end{equation}
or in element-wise form 
\begin{equation}\label{eq:differo}
\begin{split}
   \mathcal{L}_{DIFF-ERO}(E,PM_b,O_b)=
   \Sigma_{a_i \in A}\Sigma_{a_j \in A}  \\ [O_b[a_i,a_j] 
    \cdot (-\log_2 L_{PM_b}(a_i,a_j)) + (1- O_b[a_i,a_j]) \cdot (-\log_2 PM_b(a_i,a_j)) ]
\end{split}
\end{equation}

When edges are predicted to be present in $O_b$ and approach 1, they are penalized by $-log_2 L_{PM_b}$ representing the predicted distribution 
and when they are not, they are penalized by $-log_2 PM_b$ representing the ground truth. The loss computes the weighted cross-entropy between the predicted distribution $L_{PM_b}$ and the ground-truth edge distribution $PM_b$ with $O_b$ as a soft gate.
The left-hand term captures the \emph{entropy} part of the equation which aims to have a compact approximation of the ground-truth SDFA $PM_b$ using as few bits as possible. 
The right-hand term captures the \emph{relevance} part of the equation, punishing for any mismatches in SDFA structure. 
They inform the learning process about both recall and precision, contrary to, e.g., standard cross-entropy loss which is often used to optimise next-activity or suffix prediction with a focus on precision only~\cite{terven2025comprehensive}.

\subsection{\differ-Driven Topologies}
The loss function can be used in various ways which require different training procedures. We opt for a fully transformer-based architecture as they tend to return state-of-the-art performance in many tasks \cite{kappel2024attention,wuyts2024sutran}, however, our loss function is compatible with any neural network-based approach returning a hidden state, including LSTMs.
We start from a single sequence $\mathbf{x}=\langle x_1,\dots, x_l\rangle$ where each item is a one-hot encoded vector with length $|A|$. They are embedded with dimension $d$ and fed to a positional encoding layer which is processed by a multi-head self-attention block which returns an output $\mathbf{H}_{enc}\in \mathbb{R}^{|B|\times max_{l_i} \times d}$ where $max_{l_i}$ is the longest input sequence. We obtain the prefixes for all events in the training log by using $\operatorname{pref}()$ with $l=max_{l_i}-1$ set to the longest observed trace minus 1 and use padding according to Table \ref{tab:prefixes:and:suffixes}. We do the same for every event using $\operatorname{suff}()$ with the same $l$ and padding to obtain the $DF_{\sigma_{suff}}$ and $PM_b$ as described in Section \ref{sec:batches} at the output. 
In other words, we obtain a teacher forcing setup that is conformance-informed by the output's real control flow model to inform the modelling of the prefixes.

We can use the hidden state of the encoder to obtain $O$ by using a linear mapping layer over the mean-pooled version $\overline{h}$ of $\overline{\mathbf{H}}_{enc}\in \mathbb{R}^{|B|\times d}$
$$z=\overline{h} \mathbf{W}^T + b$$
which is then passed on to a softmax layer to obtain $$softmax(z)\in [0,1]^{|B|\times P}$$ where $P=|A|^2$.
Finally, we can get the soft tensor by re-structuring $softmax(z)$:
$$O\in [0,1]^{|B|\times |A|\times |A|} $$ 

\subsubsection{Auto-encoding.}
Firstly, \differ can be used as an autoencoder to perform stochastic process discovery. By learning batched prefixes with their corresponding $PM_b$ values as outputs for the \differ function, the bottleneck captures a representation which can be used, for example, for clustering or model comparison. However, given that such an unsupervised outcome is hard to verify for its usefulness, we currently did not include an experimental evaluation for this setup.

\subsubsection{Next Activity Prediction.}
Secondly, \differ can be used in a multitask setup to inform next-activity and suffix prediction. In this setup, the additional loss will inform and regularize the training process.

Then, the next-activity prediction can be informed by a joint loss function $$\mathcal{L}^{DIFF-ERO}_{na}=\mathcal{L}_{na} + \lambda\cdot \mathcal{L}_{DIFF-ERO}$$
with $\mathcal{L}_{na}= \frac{1}{|B|}\Sigma_{b=1}^{|B|} -\log(p_b(y_b))$  the batch cross-entropy loss for a vector of $|A|$ probabilities which can be used for next-activity predictions and $\lambda$ controlling the contribution of the \differ term.

\begin{figure}[htbp]
\centering
\includegraphics[width=\textwidth]{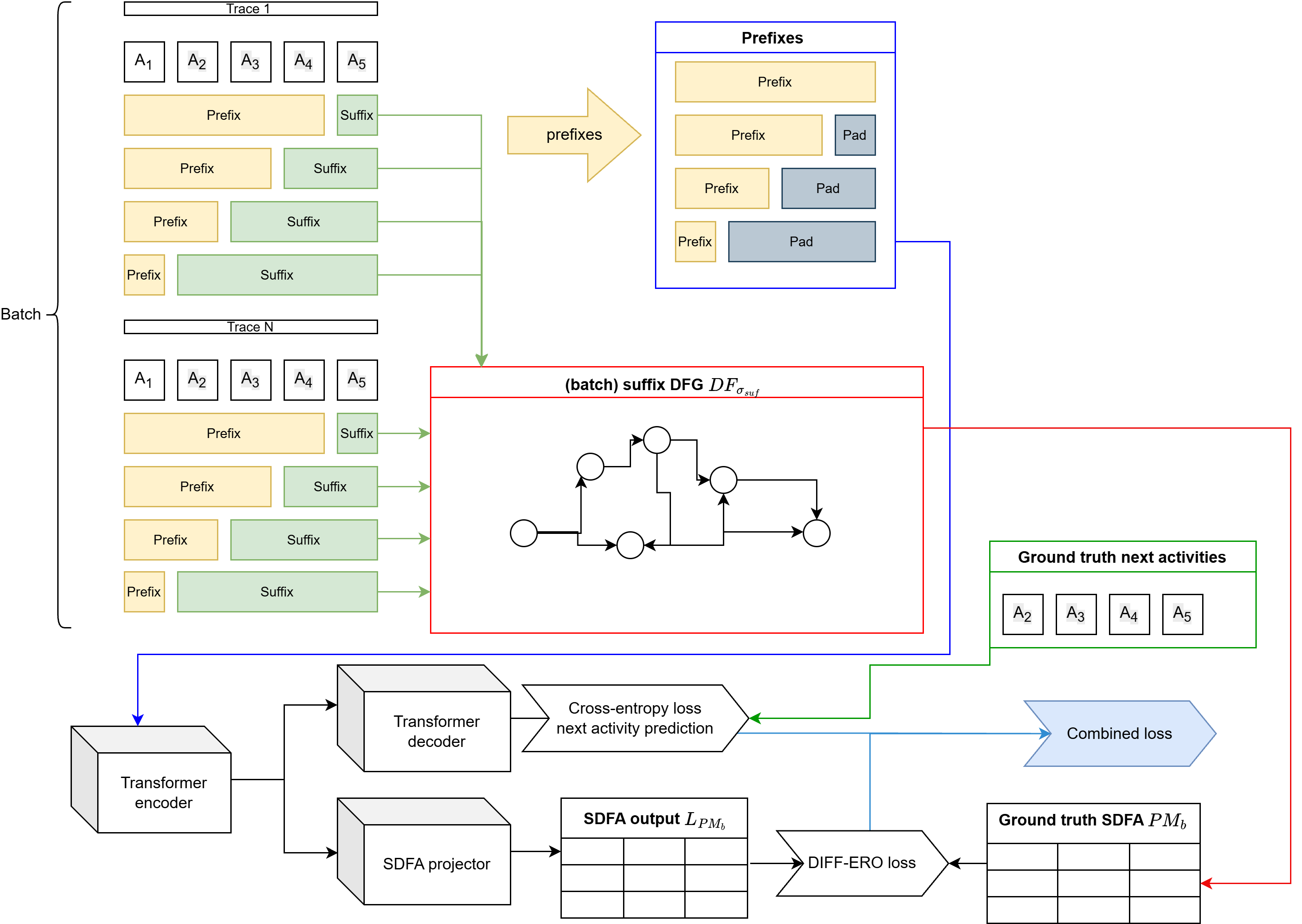}
\caption{Architectural overview of {\differ}.}
\label{fig:differ_architectur}
\end{figure}

The 
setup is illustrated in Figure \ref{fig:differ_architectur}.

\subsection{Theoretical Properties}
Below, we sketch the theoretical properties of \differ with regard to its convergence under certain distributions and the original event log-based ER.

\subsubsection{Convergence.}
We investigate the gradients to understand how $O_b$ informs the backpropagation in neural networks using $\mathcal{L}_{DIFF-ERO}(E,PM_b,O_b)$, or $\mathcal{L}$ for notational ease.

Firstly, we note that $$\frac{\partial L_{PM_b}(a_i,a_k)}{\partial O_b[a_i,a_j]}=\frac{\mathbf{1}[a_k=a_j]-L_{PM_b}(a_i,a_k)}{\Sigma_{a_3}O_b[a_i,a_3]+\epsilon}$$
for all $a_k$ sharing the row with $a_i$ which is analogous to the Jacobian of the softmax function.
Assuming that $PM_b$ is independent of $O_b$ and hence $\frac{\partial PM_b}{\partial O_b}= 0$ and knowing that $\frac{\partial}{\partial L_{PM_b}}[- log_2 L_{PM_b}]=-\frac{1}{\ln\,\, 2}\cdot\frac{1}{L_{PM_b}}$ results in the derivative  
\begin{align}\label{eq:intermediate}
 & \frac{\partial }{\partial O_b[a_i,a_j]} O_b[a_i,a_j]
    \cdot (-\log_2 L_{PM_b}(a_i,a_j))= \nonumber  \\
& -\Sigma_{a_k\in A}O_b[a_i,a_k]\cdot \frac{1}{\ln\,\, 2 \cdot L_{PM_b}(a_i,a_k)} \cdot 
\frac{\mathbf{1}[a_k=a_j]-L_{PM_b}(a_i,a_k)}{\Sigma_{a_3}O_b[a_i,a_3]+\epsilon}
\end{align}

Calculating the full derivative over 
\begin{align}
  & \frac{\partial}{\partial O_b[a_i,a_j]}
    \Bigl[
      O_b[a_i,a_j] \cdot \bigl(-\log_2 L_{PM_b}(a_i,a_j)\bigr)
      + \bigl(1 - O_b[a_i,a_j]\bigr) \cdot \bigl(-\log_2 PM_b(a_i,a_j)\bigr)
    \Bigr] \nonumber \\
  & \quad = -\log_2 L_{PM_b}(a_i,a_j) + \log_2 PM_b(a_i,a_j)  
  \quad = \log_2 \frac{PM_b(a_i,a_j)}{L_{PM_b}(a_i,a_j)}
\end{align}

and using Equations \ref{eq:lpm} and \ref{eq:intermediate} finally results in 
\begin{align}
& \frac{\partial \mathcal{L}_{DIFF-ERO}}{\partial O_b[a_i,a_j]}= \nonumber \\
& \log_2 \frac{PM_b(a_i,a_j)}{L_{PM_b}(a_i,a_j)}-\frac{1}{\ln\,\, 2 \cdot (\Sigma_{a_k} O_b[a_i,a_k]+\epsilon)}\cdot(\frac{O_b[a_i,a_j]}{L_{PM_b}(a_i,a_j)}-\Sigma_{a_k} O_b[a_i,a_k])
\end{align}

This captures three elements indicating the convergence and properties of the loss function.
Firstly, $log_2 \frac{PM_b}{L_{PM_b}}$ will align the trained $L_{PM_b}$ with the ground truth transition system in $PM_b$ by returning a positive gradient requiring further updates in the weights when $L_{PM{b}}< PM_b$ and by approaching 0 when $L_{PM_b}\simeq PM_b$.
Secondly, $O_b[a_i,a_j]$ is present as a soft gate; it reaches 0 when transitions are to be excluded thereby reducing the impact of the second term and stopping any gradients from flowing towards updating information concerning this transition. 
Finally, $\epsilon$ only appears in the denominator and avoids numerical instability by avoiding division by 0 and exploding magnitude.





\subsubsection{Consistency with ER.}
Secondly, as \differ is a soft edge membership interpretation of ER, the consistency with ER becomes clear by comparing the corresponding terms of ER and \differ.
$cost_{bits}(\sigma, E,\mathcal{A})=-log_2\,\, L_{\mathcal{A}}(\sigma)$ (Equation \ref{eq:cost}) in ER corresponds with $- \Sigma_{(a_i,a_j)\in \sigma}\,\, log_2\,\, L_{PM}(a_i,a_j)=- log_2 L_A(\sigma)$ in \differ (Equation \ref{eq:differo}). 
Hence, if the output of a parametrized model under $\theta$, $O_\theta$, converges to $PM_E$ being the SDFA of the full log, then, given that $\epsilon$ converges to 0, \differ converges to
$L_{DIFF-ERO}(E,PM_E, O_{\theta})=\frac{1}{|E|}\Sigma_{\sigma \in E} (-\log_2 L_{\mathcal{A}}(\sigma)) + C(E) = ER(E,\mathcal{A}) + C(E)$. 
Given a batch size $|b|$ equal to the event log size $|E|$ and $\epsilon=0$, \differ converges towards $ER(E,A)$ with an additive constant independent of the model's parameters.

\section{Evaluation}\label{sec5:empirical}
We have implemented the loss function in PyTorch\footnote{\url{https://anonymous.4open.science/r/differentiable-er-B7B7/README.md}, which also contains additional results including epoch time, additional statistical analysis, and full hyperparameter results}.
We next describe the experimental setup, present the next-activity prediction results, and discuss limitations.

\subsection{Experimental Setup}\label{sec51:setup}
We use six widely-used BPI Challenge event logs from of 2012, 2017, 2019, and three logs from the 2020 edition\footnote{\href{10.4121/uuid:3926db30-f712-4394-aebc-75976070e91f}{BPI12},\href{10.4121/uuid:5f3067df-f10b-45da-b98b-86ae4c7a310b}{BPI17},\href{10.4121/uuid:d06aff4b-79f0-45e6-8ec8-e19730c248f1}{BPI19},\href{10.4121/uuid:2bbf8f6a-fc50-48eb-aa9e-c4ea5ef7e8c5}{BPI20}}. Detailed statistics are provided in \Cref{tab:event_log_stats}.

\begin{table}[htbp]
\caption{Overview of the event log characteristics before and after preprocessing.}
\smallskip
\centering
\resizebox{0.8\textwidth}{!}{
  \begin{tabular}{lr *{6}{rr}}
    \toprule
    & & \multicolumn{2}{c}{\textbf{BPI12}} 
      & \multicolumn{2}{c}{\textbf{BPI17}} 
      & \multicolumn{2}{c}{\textbf{BPI19}} 
      & \multicolumn{2}{c}{\textbf{BPI20PTC}} 
      & \multicolumn{2}{c}{\textbf{BPI20RFP}} 
      & \multicolumn{2}{c}{\textbf{BPI20TPD}} \\
    \cmidrule(lr){3-4}\cmidrule(lr){5-6}\cmidrule(lr){7-8}
    \cmidrule(lr){9-10}\cmidrule(lr){11-12}\cmidrule(lr){13-14}
    \textbf{Preproc.} & & \textit{No} & \textit{Yes} 
                        & \textit{No} & \textit{Yes} 
                        & \textit{No} & \textit{Yes} 
                        & \textit{No} & \textit{Yes} 
                        & \textit{No} & \textit{Yes} 
                        & \textit{No} & \textit{Yes} \\
    \midrule
    $|A|$                    & & 24    & 26    & 26    & 28    & 42     & 38     & 29    & 31    & 19    & 20    & 51    & 53 \\
    $\overline{|\sigma|}$    & & 20.04 & 12.86 & 38.16 & 24.99 & 6.34   & 7.10   & 8.69  & 10.48 & 5.34  & 7.13  & 12.25 & 13.21 \\
    $\text{median}(|\sigma|)$& & 11    & 8     & 35    & 23    & 5      & 7      & 8     & 10    & 5     & 7     & 11    & 12 \\
    $|C|$                    & & 13{,}087 & 12{,}595 & 31{,}509 & 30{,}189 & 251{,}734 & 239{,}734 & 2{,}099 & 2{,}030 & 6{,}886 & 6{,}579 & 7{,}065 & 6{,}714 \\
    $\max(|\sigma|)$          & & 175   & 35    & 180   & 41    & 990    & 10     & 21    & 15    & 20    & 10    & 90    & 23 \\
    \bottomrule
  \end{tabular}
}
\label{tab:event_log_stats}
\end{table}

To streamline the predictive task, we retain traces within the 95\% trace length distribution and 
apply out-of-time correction to prevent temporal leakage~\cite{weytjens2021creating}.
This results in an 80/20\% train-test split without temporal overlap.
The prefix input-length $l$ is set to the maximum observed training prefix length. 
Transformer dimensionality (encoder, decoder, and $PM_b$ projection layer) is varied across 16, 32, and 64, while the number of heads and layers is fixed at 4 and 2 following \cite{kappel2024attention}.
For \differ, we evaluate local batch-level $PM_b$ tensors instead of training-wide $PM$ tensors to study the effect of regularisation and varying SDFA granularity. Since batch size influences the discovered variants, we vary it between 16, 32, and 128, similarly to imputing different discovered models~\cite{mezini5420992neuro}. Our focus, however, is on the effect of the loss function rather than SDFA discovery quality.

Finally, we vary $\lambda$ controlling the contribution of the entropic process loss.
Each configuration is evaluated over 10 random seeds for statistical analysis.

For next-activity prediction, we report the weighted precision and F1-score as the harmonic mean of recall and precision of predicting the next activity in the trace. The weighing corrects the macro-average for class imbalance, which is appropriate given the skewed activity frequency distribution.
Finally, we also visualise the loss over training to assess whether the model is actually informed by the loss during training, and acts on it during backpropagation, i.e., whether the values for $\mathcal{L}_{DIFF-ERO}$ decrease.

We compare \differ to other loss and target baselines.
\begin{enumerate}
    \item Base model (\texttt{base}): we start from a baseline encoder-decoder transformer-based model using cross-entropy for training because of its strong performance for next-activity prediction \cite{kappel2024attention}. Note that all models below can also be used directly with other neural network architectures like LSTMs.
    \item Base model with SDFA (\texttt{SDFA}): to verify whether adding the extra information of the ground truth SDFA can already inform the training of the network without dedicated structural loss, we add the SDFA to the output of the model and backpropagate using standard cross-entropy loss. This can happen both in a batch form and a global form as explained in Section \ref{sec:batches}.
    \item Base model with rank loss (\texttt{rank\_loss}): in order to validate whether altering the loss in the form of another information retrieval baseline can improve the performance, we employ the rank loss \cite{chen2009ranking} which performs negative sampling to understand whether a learning-to-rank baseline already can improve a cross-entropy optimisation while still not directly incorporating a process structural point of view. The rank loss is $L_b=\frac{1}{|\mathcal{N}_b|} \Sigma_{a_j \in \mathcal{N}_b} \log (1+ e^{logit(a_j)-logit(y_b)}$ where $\mathcal{N}_b$ is the number of negative samples and $logit()$ is the logit scores of these samples.
    \item Base model with \differ (\texttt{DIFF-ERO}): finally, we use the base model with the full \differ loss to leverage the SDFA information as an additional target in both local/batch and global form.
\end{enumerate}


\subsection{Results}\label{sec52:results}
The results for next-activity prediction (with optimal hyperparameters), averaged over random seeds, are shown in Table \ref{tab:nap_results}. \differ-based models report the best or among the best results in terms of F1-score and are comparable in terms of precision, suggesting stronger recall. This further underlines that it steers the model in additional ways compared to standard cross-entropy-based models. Still, the strong performance of the \texttt{SDFA}-based models, with the exception of BPI12, indicates that adding the additional data at the output of the model is a major driver of this effect. Rank loss does not seem to improve training and rather reports worse results compared to the baseline. Based on training time per epoch, it is clear that \differ adds only a modest overhead compared to rank loss or incorporating SDFAs using cross-entropy.
\vspace{-7mm}

\begin{table}[htbp]
  \centering
\caption{Results of the next activity prediction tests with best performances in boldface.}
 \resizebox{\textwidth}{!}{
    \begin{tabular}{|l|cccc|cccc|cccc|}
\cline{2-13}    \multicolumn{1}{r|}{} & \multicolumn{4}{c|}{\textbf{F1}} & \multicolumn{4}{c|}{\textbf{Precision}} & \multicolumn{4}{c|}{\textbf{Time (seconds per epoch)}} \bigstrut\\
    \hline
    \textbf{Dataset} & \multicolumn{1}{c}{\textbf{Base}} & \multicolumn{1}{c}{\textbf{Rank loss}} & \multicolumn{1}{c}{\textbf{SDFA}} & \multicolumn{1}{c|}{\textbf{DIFF-ERO}} & \multicolumn{1}{c}{\textbf{Base}} & \multicolumn{1}{c}{\textbf{Rank loss}} & \multicolumn{1}{c}{\textbf{SDFA}} & \multicolumn{1}{c|}{\textbf{DIFF-ERO}} & \multicolumn{1}{c}{\textbf{Base}} & \multicolumn{1}{c}{\textbf{Rank loss}} & \multicolumn{1}{c}{\textbf{SDFA}} & \multicolumn{1}{c|}{\textbf{DIFF-ERO}} \bigstrut\\
    \hline
    \textbf{BPI12} & 0.786 & 0.716 & 0.715 & \textbf{0.826} & 0.878 & 0.83  & 0.863 & \textbf{0.882} & 19.447 & 49.41 & 25.528 & 19.709 \bigstrut\\
    \textbf{BPI17} & 0.823 & 0.795 & 0.83  & \textbf{0.865} & 0.895 & 0.883 & 0.899 & \textbf{0.901} & 136.267 & 250.709 & 144.111 & 144.7 \\
    \textbf{BPI19} & 0.722 & 0.735 & 0.769 & \textbf{0.79} & 0.918 & 0.914 & 0.918 & 0.916 & 85.574 & 293.725 & 113.466 & 88.273 \\
    \textbf{BPI20PTC} & 0.855 & 0.818 & 0.887 & \textbf{0.89} & 0.939 & 0.936 & \textbf{0.955} & 0.944 & 2.323 & 7.17  & 2.922 & 2.408 \\
    \textbf{BPI20RFP} & 0.948 & 0.948 & 0.949 & 0.949 & 0.99  & 0.989 & 0.99  & 0.99  & 4.343 & 15.415 & 5.474 & 4.558 \\
    \textbf{BPI20TPD} & 0.808 & 0.737 & 0.854 & \textbf{0.862} & 0.906 & 0.884 & \textbf{0.926} & \textbf{0.924} & 11.46 & 32.429 & 17.736 & 12.13 \bigstrut\\
    \hline
    \end{tabular}%
   }
  \label{tab:nap_results}%
\end{table}%

\vspace{-3mm}

\begin{figure}[htbp]
\centering
\begin{minipage}[c]{0.48\textwidth}
  \centering
  \includegraphics[width=\textwidth]{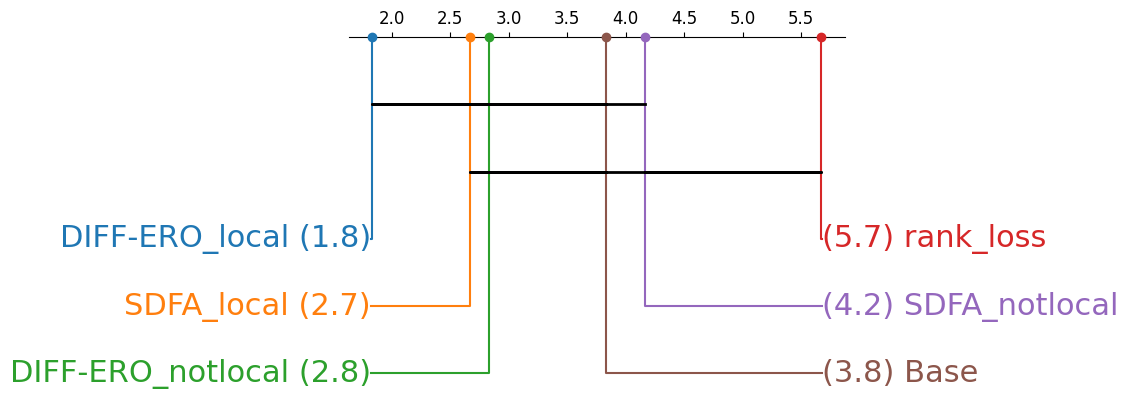}
  \subcaption{Nemenyi post-hoc test of NAP results.}
  \label{fig:nemenyi}
\end{minipage}
\hfill
\begin{minipage}[c]{0.48\textwidth}
  \centering
  \resizebox{\textwidth}{!}{
    \begin{tabular}{lrrrrr}
      \toprule
      & \textbf{(2)} & \textbf{(3)} & \textbf{(4)} & \textbf{(5)} & \textbf{(6)} \\
      \midrule
      (1) Base                             & 0.432 & 0.940 & 0.889 & 1.000 & 0.534 \\
      (2) DIFF-ERO\textsubscript{local}    &       & 0.940 & 0.972 & 0.257 & 0.005 \\
      (3) DIFF-ERO\textsubscript{notlocal} &       &       & 1.000 & 0.820 & 0.092 \\
      (4) SDFA\textsubscript{local}        &       &       &       & 0.734 & 0.061 \\
      (5) SDFA\textsubscript{notlocal}     &       &       &       &       & 0.734 \\
      \bottomrule
    \end{tabular}
  }
  \subcaption{Pairwise $p$-values between methods.}
  \label{tab:pvalues}
\end{minipage}

\caption{Nemenyi post-hoc results and corresponding pairwise $p$-values of the F1-score.}
\label{fig:nemenyi_combined}
\end{figure}

A rank-based Friedman test is used to compare the significance in the results of each model over the different datasets where a consistently lower rank indicates a better performing model. The Nemenyi test is a post-hoc test which reports which pair-wise comparisons are significantly different. The average rank of each technique (lower is better) and the corresponding Nemenyi outcome is visualised in Figure \ref{fig:nemenyi}.

\begin{figure}[H]
    \centering

    \begin{subfigure}[t]{0.30\textwidth} 
        \centering
        \includegraphics[width=\linewidth, trim = 15mm 5mm 15mm 0mm]{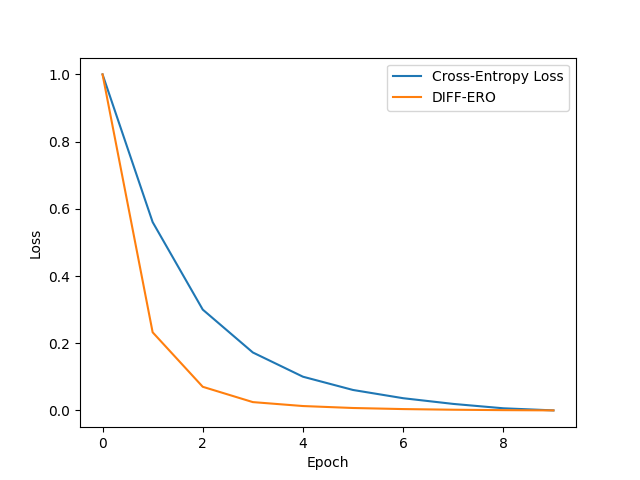} 
        \caption{BPI12}
        \label{fig:sub1}
    \end{subfigure}
    \hfill
    \begin{subfigure}[t]{0.30\textwidth}
        \centering
        \includegraphics[width=\linewidth, trim = 15mm 5mm 15mm 0mm]{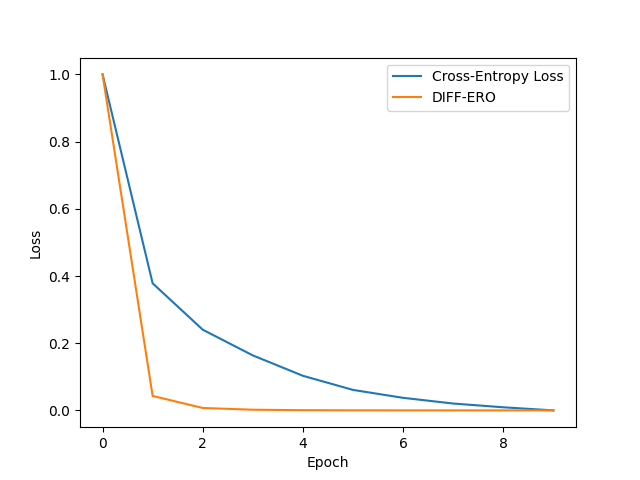} 
        \caption{BPI17}
        \label{fig:sub2}
    \end{subfigure}
    \hfill
    \begin{subfigure}[t]{0.30\textwidth}
        \centering
        \includegraphics[width=\linewidth, trim = 15mm 5mm 15mm 0mm]{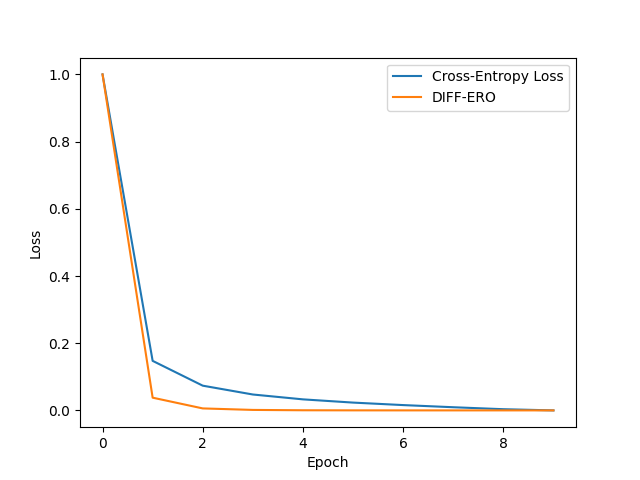} 
        \caption{BPI19}
        \label{fig:sub3}
    \end{subfigure}
    \caption{Evolution of training loss of cross-entropy and \differ~for next activity 
    prediction.}
    \label{fig:loss_over_epoch}
\end{figure}

\differ does not outperform the \texttt{SDFA} and \texttt{base} models at a significance level of 95\% due to the relatively small number of datasets and relatively strong performance of all models for BPI20PTC and BPI20RFP.
While DIFF-ERO does not statistically outperform the baselines on pure predictive accuracy due to the use of a limit number of datasets, it achieves the highest overall rank while guaranteeing structural convergence (as seen in Figure \ref{fig:loss_over_epoch}) while incurring only modest additional computational cost. This demonstrates that the network internalizes global process constraints without sacrificing local predictive power.

A closer examination shows that non-local models, typically with higher $\lambda$ values and batch sizes, achieve the best performance. Unlike standard cross-entropy, \differ uses batch size as a structural regularizer: larger batches provide a more complete representation of the ground-truth SDFA, increasing the emphasis on structural recall through the left-hand term of Equation~\ref{eq:differo}.
Table~\ref{tab:nap_results} further shows that \differ is less sensitive to the global-local setup than \texttt{SDFA}, where global SDFAs generally perform worse. In this sense, the \differ construction appears more robust.
Finally, Figure~\ref{fig:loss_over_epoch} presents the min-max scaled cross-entropy loss and \differ over training epochs for the best-performing configurations on the BPI12/17/19 logs. All models show a clear and rapidly converging decrease in \differ alongside the standard cross-entropy loss, particularly for the larger BPI17 and BPI19 logs. This indicates that the model successfully propagates the structural signal from the SDFA through its hidden layers during training.
\vspace{-10mm}
\begin{table}
\caption{Detailed F1-score results with non-local (top 4 rows for standard batch size 32) and local models with different batch sizes. Higher values are in red, lower values in blue. Boldface indicates best performance.}
\smallskip
  \centering
\resizebox{.75\textwidth}{!}{
    \begin{tabular}{crrrrrrr}
    \toprule
    \multicolumn{1}{l}{\textbf{$|b|$}} & \multicolumn{1}{l}{\textbf{lambda}} & \multicolumn{1}{l}{\,\,\,\,\,\,\textbf{BPI12}\,\,\,} & \multicolumn{1}{l}{\,\,\,\,\,\,\textbf{BPI17}\,\,\,} & \multicolumn{1}{l}{\,\,\,\,\,\,\textbf{BPI19}\,\,\,} & \multicolumn{1}{l}{\textbf{BPI20PTC}} & \multicolumn{1}{l}{\textbf{BPI20RFP}} & \multicolumn{1}{l}{\textbf{BPI20TPD}} \\
    \midrule
    \multirow{4}[2]{*}{\textbf{32}} & \textbf{0} & \cellcolor[rgb]{ .827,  .875,  .941}0.769 & \cellcolor[rgb]{ .671,  .765,  .886}0.800 & \cellcolor[rgb]{ .988,  .878,  .886}0.690 & \cellcolor[rgb]{ .98,  .627,  .635}0.812 & \cellcolor[rgb]{ .98,  .663,  .671}0.948 & \cellcolor[rgb]{ .353,  .541,  .776}0.779 \\
          & \textbf{0.1} & \cellcolor[rgb]{ .984,  .843,  .855}0.778 & \cellcolor[rgb]{ .725,  .804,  .906}0.801 & \cellcolor[rgb]{ .925,  .945,  .976}0.685 & \cellcolor[rgb]{ .984,  .741,  .749}0.806 & \cellcolor[rgb]{ .98,  .576,  .588}0.948 & \cellcolor[rgb]{ .522,  .659,  .835}0.784 \\
          & \textbf{0.2} & \cellcolor[rgb]{ .98,  .6,  .608}0.783 & \cellcolor[rgb]{ .808,  .863,  .937}0.804 & \cellcolor[rgb]{ .976,  .502,  .51}0.700 & \cellcolor[rgb]{ .988,  .988,  1}0.795 & \cellcolor[rgb]{ .988,  .988,  1}0.947 & \cellcolor[rgb]{ .976,  .498,  .51}0.812 \\
          & \textbf{0.5} & \cellcolor[rgb]{ .976,  .424,  .431}0.788 & \cellcolor[rgb]{ .976,  .478,  .49}0.819 & \cellcolor[rgb]{ .984,  .788,  .796}0.692 & \cellcolor[rgb]{ .424,  .588,  .8}0.709 & \cellcolor[rgb]{ .463,  .616,  .812}0.943 & \cellcolor[rgb]{ .843,  .886,  .949}0.795 \\
    \midrule
    \multirow{3}[2]{*}{\textbf{16}} & \textbf{0.1} & \cellcolor[rgb]{ .953,  .961,  .984}0.773 & \cellcolor[rgb]{ .353,  .541,  .776}0.790 & \cellcolor[rgb]{ .988,  .965,  .976}0.688 & \cellcolor[rgb]{ .98,  .573,  .58}0.814 & \cellcolor[rgb]{ .988,  .871,  .882}0.947 & \cellcolor[rgb]{ .8,  .855,  .933}0.793 \\
          & \textbf{0.2} & \cellcolor[rgb]{ .988,  .882,  .89}0.777 & \cellcolor[rgb]{ .651,  .753,  .882}0.799 & \cellcolor[rgb]{ .973,  .412,  .42}\textbf{0.702} & \cellcolor[rgb]{ .941,  .957,  .984}0.788 & \cellcolor[rgb]{ .898,  .922,  .965}0.946 & \cellcolor[rgb]{ .988,  .918,  .929}0.801 \\
          & \textbf{0.5} & \cellcolor[rgb]{ .976,  .525,  .537}0.785 & \cellcolor[rgb]{ .988,  .914,  .925}0.811 & \cellcolor[rgb]{ .78,  .843,  .925}0.679 & \cellcolor[rgb]{ .353,  .541,  .776}0.699 & \cellcolor[rgb]{ .353,  .541,  .776}0.943 & \cellcolor[rgb]{ .988,  .894,  .906}0.802 \\
    \midrule
    \multirow{3}[2]{*}{\textbf{32}} & \textbf{0.1} & \cellcolor[rgb]{ .847,  .89,  .949}0.769 & \cellcolor[rgb]{ .988,  .906,  .918}0.811 & \cellcolor[rgb]{ .953,  .961,  .984}0.686 & \cellcolor[rgb]{ .976,  .447,  .455}0.820 & \cellcolor[rgb]{ .984,  .733,  .741}0.947 & \cellcolor[rgb]{ .392,  .569,  .788}0.780 \\
          & \textbf{0.2} & \cellcolor[rgb]{ .973,  .976,  .992}0.774 & \cellcolor[rgb]{ .973,  .412,  .42}\textbf{0.821} & \cellcolor[rgb]{ .808,  .859,  .933}0.680 & \cellcolor[rgb]{ .984,  .984,  .996}0.794 & \cellcolor[rgb]{ .984,  .984,  .996}0.947 & \cellcolor[rgb]{ .976,  .439,  .447}0.813 \\
          & \textbf{0.5} & \cellcolor[rgb]{ .957,  .965,  .988}0.773 & \cellcolor[rgb]{ .984,  .804,  .816}0.813 & \cellcolor[rgb]{ .969,  .973,  .992}0.686 & \cellcolor[rgb]{ .416,  .584,  .796}0.708 & \cellcolor[rgb]{ .494,  .639,  .824}0.944 & \cellcolor[rgb]{ .988,  .91,  .922}0.801 \\
    \midrule
    \multirow{3}[2]{*}{\textbf{64}} & \textbf{0.1} & \cellcolor[rgb]{ .353,  .541,  .776}0.752 & \cellcolor[rgb]{ .937,  .953,  .98}0.808 & \cellcolor[rgb]{ .98,  .682,  .694}0.695 & \cellcolor[rgb]{ .973,  .412,  .42}\textbf{0.822} & \cellcolor[rgb]{ .973,  .412,  .42}\textbf{0.948} & \cellcolor[rgb]{ .675,  .765,  .886}0.789 \\
          & \textbf{0.2} & \cellcolor[rgb]{ .851,  .89,  .949}0.770 & \cellcolor[rgb]{ .984,  .788,  .8}0.813 & \cellcolor[rgb]{ .78,  .843,  .925}0.679 & \cellcolor[rgb]{ .976,  .545,  .553}0.816 & \cellcolor[rgb]{ .878,  .91,  .961}0.946 & \cellcolor[rgb]{ .773,  .835,  .922}0.792 \\
          & \textbf{0.5} & \cellcolor[rgb]{ .973,  .412,  .42}\textbf{0.788} & \cellcolor[rgb]{ .976,  .553,  .565}0.818 & \cellcolor[rgb]{ .988,  .949,  .961}0.688 & \cellcolor[rgb]{ .576,  .698,  .855}0.733 & \cellcolor[rgb]{ .533,  .667,  .839}0.944 & \cellcolor[rgb]{ .976,  .533,  .541}0.811 \\
    \midrule
    \multirow{3}[2]{*}{\textbf{128}} & \textbf{0.1} & \cellcolor[rgb]{ .98,  .631,  .639}0.783 & \cellcolor[rgb]{ .922,  .941,  .976}0.808 & \cellcolor[rgb]{ .98,  .573,  .584}0.698 & \cellcolor[rgb]{ .988,  .847,  .859}0.801 & \cellcolor[rgb]{ .988,  .851,  .863}0.947 & \cellcolor[rgb]{ .929,  .945,  .976}0.797 \\
          & \textbf{0.2} & \cellcolor[rgb]{ .753,  .824,  .918}0.766 & \cellcolor[rgb]{ .714,  .796,  .902}0.801 & \cellcolor[rgb]{ .353,  .541,  .776}0.662 & \cellcolor[rgb]{ .953,  .965,  .988}0.790 & \cellcolor[rgb]{ .988,  .89,  .902}0.947 & \cellcolor[rgb]{ .984,  .816,  .824}0.804 \\
          & \textbf{0.5} & \cellcolor[rgb]{ .988,  .969,  .98}0.775 & \cellcolor[rgb]{ .988,  .863,  .871}0.812 & \cellcolor[rgb]{ .573,  .698,  .855}0.671 & \cellcolor[rgb]{ .459,  .616,  .812}0.715 & \cellcolor[rgb]{ .451,  .608,  .808}0.943 & \cellcolor[rgb]{ .973,  .412,  .42}\textbf{0.814} \\
    \bottomrule
    \end{tabular}
    }%
  \label{tab:hyper_nap}%
\end{table}%
\vspace{-15mm}

\section{Discussion and Limitations}
\label{sec:discussion}
\differ\ introduces a differentiable mechanism to incorporate conformance information directly into neural training, enabling models to learn structures that more closely reflect process behaviour. 
Currently, the approach is operationalized as a regularization for next activity prediction with similar aims like the rank-based baselines.
Contrary to prior approaches incorporating conformance information \cite{donadello2023knowledge,mezini5420992neuro}, it offers a trade-off between inputting a coarser, continuous stochastic automaton to incorporate conformance information versus a more fine-grained constraint-based model or procedural model \cite{donadello2023knowledge,umili2023grounding} which require approximations to obtain differentiability over discrete domains.
Imputation currently happens through batches of the training set, however, similar to \cite{mezini5420992neuro,stevens2025generating,umili2023grounding}, an additional discovery step could further lead \differ to be used to input specific information in the form of a ground truth $PM_b$ external to the (training) data. \Cref{tab:hyper_nap} illustrates the impact of changing the model used in the loss function.

Our experiments show significant improvements in next-activity prediction. 
These findings suggest that the proposed loss function may also be useful beyond prediction tasks, particularly in settings where adherence to process structure is essential.
However, we have not yet evaluated additional predictive tasks that could benefit from the structural information. Most notably, it would be interesting to use \differ in the context of suffix prediction, where the state information infused into the training process could improve the generation of next activities that adhere to a process model.

At the same time, several limitations accompany these contributions. The composite loss introduces computational overhead that remains modest for small and medium alphabets. Still, it may increase memory usage and training time for logs with large activity sets or highly imbalanced distributions. The weighting parameter~$\lambda$ offers flexibility but adds an additional hyperparameter, possibly requiring tuning. Moreover, the current formulation captures only pairwise directly-follows relations, limiting its ability to express higher-order or concurrent behaviour such as loops, parallelism, and so on, without further extension.
The results in Table \ref{tab:hyper_nap} and Figure \ref{fig:loss_over_epoch} also suggest that a further study into the relation between convergence, batch size, and even learning rate could further clarify the impact of \differ.
Secondly, the grouping within batches was not explored in detail. Clustering-based approaches to group prefixes could potentially further leverage the power of \differ.
Finally, the experiments focus exclusively on predictive models that rely on control-flow information, excluding payload or contextual data. Although the inclusion of such data can improve predictive performance~\cite{wuyts2024sutran}, it was intentionally omitted here, given the nature of the experiments where the focus lies on demonstrating its potential use for learning control-flow behaviour.

\section{Conclusion}\label{sec:conclusion}
This paper introduced a process model-informed loss function for deep learning on event logs that exposes structural process model features to backpropagation. 
The entropy-based process loss {\differ} proposed is fully differentiable, can be computed at batch level, and is compatible with architectures that parametrise stochastic transitions. We instantiated \differ in a transformer encoder-decoder architecture for next-activity prediction.
Our empirical results show that {\differ} significantly improves next-activity prediction accuracy when compared to a cross-entropy baseline. In both tasks, the learned representations converge towards the structural ground truth, leading to models whose internal stochastic automata better reflect the underlying process. This suggests that {\differ} supports structurally faithful representation learning, even when predictive gains are limited. 
For future work, we envision applying {\differ} in other deep learning contexts, including auto-encoding for unsupervised tasks and prescriptive settings in which predictive models are used to generate generalizable recommendations to optimise process performance.



\bibliographystyle{splncs04}

\end{document}